\documentclass[final,5p,times,authoryear]{elsarticle}

\usepackage{amssymb}
\usepackage{amsmath}
\usepackage{amsthm}
\usepackage{hyperref}
\usepackage{booktabs}
\usepackage{multirow}
\usepackage{subcaption}
\usepackage{arydshln}
\usepackage{graphicx}
\usepackage[table,xcdraw]{xcolor}
\usepackage[normalem]{ulem}
\useunder{\uline}{\ul}{}
\usepackage{algorithm}
\usepackage{algpseudocode}

\newtheorem{theorem}{Theorem}[section]

\newtheorem{lemma}[theorem]{Lemma}

\theoremstyle{definition}

\journal{}

\begin{document}

\begin{frontmatter}



\title{Modality as Heterogeneity: Node Splitting and Graph Rewiring for \\Multimodal Graph Learning} 


\author[1]{Yihan Zhang} 
\ead{zyh23@mails.tsinghua.edu.cn}

\author[1]{Ercan E. Kuruoglu\corref{cor1}} 
\ead{kuruoglu@sz.tsinghua.edu.cn}
\cortext[cor1]{Corresponding author.}

\affiliation[1]{organization={Institute of Data and Information, Shenzhen International Graduate School, Tsinghua University},
            city={Shenzhen},
            country={China}}

\begin{abstract}
Multimodal graphs are gaining increasing attention due to their rich representational power and wide applicability, yet they introduce substantial challenges arising from severe modality confusion.
To address this issue, we propose NSG (Node Splitting Graph)-MoE, a multimodal graph learning framework that integrates a node-splitting and graph-rewiring mechanism with a structured Mixture-of-Experts (MoE) architecture.
It explicitly decomposes each node into modality-specific components and assigns relation-aware experts to process heterogeneous message flows, thereby preserving structural information and multimodal semantics while mitigating the undesirable mixing effects commonly observed in general-purpose GNNs.
Extensive experiments on three multimodal benchmarks demonstrate that NSG-MoE consistently surpasses strong baselines.
Despite incorporating MoE—which is typically computationally heavy—our method achieves competitive training efficiency.
Beyond empirical results, we provide a spectral analysis revealing that NSG performs adaptive filtering over modality-specific subspaces, thus explaining its disentangling behavior.
Furthermore, an information-theoretic analysis shows that the architectural constraints imposed by NSG reduces mutual information between data and parameters and improving generalization capability.
\end{abstract}




\end{frontmatter}



\section{Introduction}

Graphs provide a powerful and flexible abstraction for modeling relational data in numerous real-world applications, including knowledge management, e-commerce, biomedicine, and social networks. Recent progress in text-attributed graph learning \citep{yan2023comprehensive} demonstrates the importance of enriching graph nodes with semantic information, such as descriptions or documents, enabling models to leverage textual context alongside structural patterns. However, many real systems contain far richer multimodal signals: a webpage incorporates visual, textual, and sometimes audio content \citep{burns2023a}; and clinical records combine imaging, lab measurements, and reports \citep{zheng2022multi}. This motivates multimodal graph learning \citep{ektefaie2023multimodal}, which seeks to jointly model graph structure and multimodal content to learn more expressive node representations and more accurate predictions.
Multimodal data, however, poses fundamental challenges. Modalities differ significantly in structure, embedding space, semantic granularity, and noise characteristics. Mainstream multimodal transformers \citep{lu2019vilbert,tan2019lxmert} and contrastive encoders such as CLIP \citep{radford2021learning} and ImageBind \citep{girdhar2023imagebind} provide powerful cross-modal alignment, yet integrating such features into graph neural networks (GNNs) remains non-trivial.
Multimodal graph learning requires not only the fusion of heterogeneous feature spaces but also propagating them through relational neighborhoods—where naïve fusion often causes modality confusion, a collapse of modality-specific patterns during message passing.

Current interpretations of multimodal graphs vary. Some studies focus on cases where nodes in the graph contains information of only a single modality \citep{chen2022hybrid,yoon2023multimodal}, whereas others treat nodes as inherently multimodal and rely on upstream encoders to merge modalities \citep{zhu2025mosaic,he2025unigraph2}.
In fact, the first case is encompassed within the second case, and can be viewed as some modalities being masked. Therefore, in this paper, we will only discuss the more general second case.
The prevalent approach concatenates modality-specific embeddings into a single node feature vector before applying a GNN. While simple, this practice ignores the fact that textual, visual, and other modalities inhabit meaningfully different embedding spaces—even after alignment by pretrained models—and GNNs are not inherently designed to disentangle such spaces. Consequently, standard message passing tends to blur modality-dependent semantics, reducing expressive power and limiting generalization.

\begin{figure}[t]
    \centering
    \includegraphics[width=\linewidth]{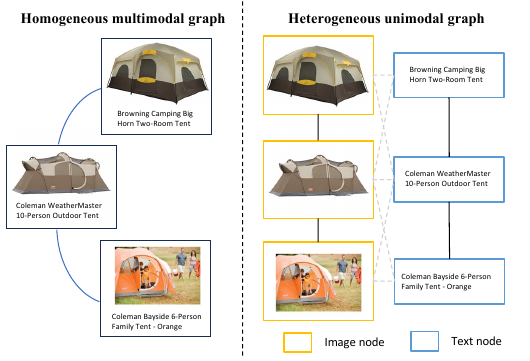}
    \caption{Intuitive understanding of the proposed node splitting mechanism. Different colors represent different modalities. Data source: \citep{zhu2025mosaic}.}
    \label{fig:examp}
\end{figure}

In this work, we argue that multimodal nodes should not be treated as homogeneous entities. Instead, we propose to explicitly model modalities as heterogeneous components within the graph.
To achieve this, we introduce a node-splitting mechanism (Figure \ref{fig:examp}) that decomposes each multimodal node into multiple unimodal nodes, one per modality. We then reconstruct the graph by preserving its original topology while enabling structured interactions between the split nodes. This treatment transforms multimodal graphs into a form of heterogeneous graphs, but with heterogeneity rooted in modal semantics rather than manually assigned node types.
Earlier multimodal recommendation models \citep{wei2019mmgcn,zhang2021mining} hinted at separating modalities during propagation, but subsequent works largely returned to fusion-first strategies. MMGAT \citep{zhang2025moe} reintroduces modality separation but still applies homogeneous message passing to each modality independently, limiting cross-modal relational reasoning. In contrast, our formulation leverages heterogeneous graph analysis to guide both intra-modal and cross-modal propagation while avoiding the pitfalls of naïve fusion.
Furthermore, multimodal graphs can grow rapidly in scale as the number of modalities increases. To maintain efficiency without degrading representational capacity, we propose a modality-aware sparsification method based on maximum spanning tree algorithm that selectively preserves high-utility connections.

To better understand the behavior of the proposed framework, we provide theoretical analysis by reducing the heterogeneous GNN operations into a more general linear form and conducting spectral investigations. This reveals that the node-splitting architecture induces disentangled low-pass filters across modality-specific subspaces, allowing adaptive smoothing behavior that differs for each modality. We also derive a generalization bound by showing that our model constitutes a structured instance of GNNs.

Beyond the structural redesign, we enhance model expressiveness using a Graph Mixture of Experts (GraphMoE) \citep{wang2023graph} module. Each expert is implemented as a heterogeneous graph neural network (HGNN) and each expert group specializes in a distinct relational pattern. The gating mechanism adaptively route structural information to different experts, ensuring that each node receives messages from the neighbors that contribute the most to it. Unlike previous work, which employs MLP-based experts focused solely on node-level transformation \citep{he2025unigraph2} or GNN-based experts that performs homogeneous message passing \citep{zhang2025moe}, our design uses HGNN experts capable of learning specialized heterogeneous subgraph behaviors. This facilitates multi-view reasoning across modalities and allows experts to capture distinct structural and semantic patterns that arise from node splitting and graph rewiring.

Our main contributions are as follows:
\begin{itemize}
    \item We propose a novel method for multimodal graph learning, which handles modality as heterogeneity by splitting multimodal nodes into multiple unimodal nodes and rewiring graph. It enables structured intra- and cross-modal message passing and scales effectively when the number of modalities is large.
    \item We provide theoretical analyses by analyzing heterogeneous GNN operators in the spectral domain, showing that our model performs adaptive low-pass filtering in decomposed subspaces. We also derive a lower generalization bound of our model through information-theoretic analysis.
    \item We extend the basic learning method by leveraging GraphMoE framework in an intuitive way. The designed two categories of experts jointly capture cross-modal correlations and within-modality variations, with multiple experts per category improving the ability to represent diverse multimodal distributions.
    \item Extensive experiments on two multimodal graph benchmarks and one independent dataset across node classification and link prediction demonstrate consistent improvements over state-of-the-art baselines, with further ablations verifying the role of each module and efficiency studies confirming scalability.
\end{itemize}

The remainder of the paper is organized as follows. Section 2 reviews related work. Section 3 introduces preliminaries. Section 4 presents the proposed node-splitting framework. Section 5 offers theoretical analysis. Section 6 introduces the GraphMoE extension. Section 7 reports experimental results, and Section 8 concludes the study.

\section{Related Work}
\subsection{Multimodal Graph Learning}
Multimodal representation learning seeks to embed data from diverse modalities—such as text, images, audio, and sensor signals—into a shared semantic space \citep{radford2021learning,girdhar2023imagebind}. Multimodal graph learning extends this objective to graph-structured data \citep{ektefaie2023multimodal,zhu2025mosaic,yan2025graph}, where nodes are associated with heterogeneous modal features. The core challenge is to capture both the structural dependencies intrinsic to the graph and the semantic complementarities across modalities.
Most early research on multimodal graphs has concentrated on recommendation systems \citep{wei2019mmgcn,guo2024lgmrec,yu2025mind}, knowledge graph completion \citep{chen2022hybrid}, and multimodal generative tasks \citep{yoon2023multimodal}. While effective in their specific domains, these models typically rely on narrow task assumptions—such as user–item bipartite structures or predefined relational schemas—which limits their transferability to more general multimodal graph settings.
To address the need for universal frameworks, several recent works attempt to build foundation-style multimodal graph models. MMGL \citep{zheng2022multi} introduces general multimodal graph formulations, and Unigraph2 \citep{he2025unigraph2} further moves toward unifying multimodal alignment and graph reasoning. However, these systems predominantly adopt early-fusion strategies: multimodal features are fused at the node level prior to message passing, without explicitly modeling the multi-level correlations between modalities across nodes. Consequently, they struggle to disentangle modality-specific effects or exploit cross-modal relational patterns embedded in the graph.
This motivates the need for frameworks that integrate modality-aware graph restructuring with richer cross-type relational modeling.

\subsection{Heterogeneous Graphs}
A system composed of multi-typed and interconnected entities can be modeled as heterogeneous graphs (HGs), also known as heterogeneous information networks. Unlike homogeneous graphs, which contain only one type of node and edge, HGs explicitly preserve the rich semantic information embedded in different object and relation types. Correspondingly, Heterogeneous Graph Neural Networks (HGNNs) have been proposed to process HGs through type-specific transformations and relation-dependent message passing \citep{wang2019heterogeneous,hu2020heterogeneous}.
Several studies explore applying HGNNs to multimodal learning by treating multimodal links as heterogeneous relations \citep{chen2020hgmf,kim2023heterogeneous}. Although this approach leverages coarse-grained heterogeneity in edges, it assumes that all nodes share the same type. As a result, modality-level distinctions within nodes remain entangled, and the model may conflate fundamentally different semantic sources.
Addressing modality confusion thus requires a more fine-grained perspective than traditional HGNN formulations provide.

\subsection{Graph Mixture of Experts}
The Mixture of Experts (MoE) paradigm improves model capacity and specialization by distributing computation across multiple expert networks and routing each sample to the most relevant subset of experts through a gating function \citep{shazeer2017}.
Sparse activation makes MoE computationally efficient, while the dynamic routing mechanism encourages experts to specialize in different patterns or subspaces.
Graph Mixture of Experts (GMoE) extends MoE to graph learning by replacing the standard feed-forward experts with GNN-based experts \citep{wang2023graph}. Each expert thus becomes specialized not only in feature transformation but also in distinct structural patterns or receptive fields—such as 1-hop versus 2-hop neighborhoods—allowing the model to allocate computation based on local graph structure.
Subsequent works refine this principle by modifying the aggregation functions \citep{guo2025graphmore} or adopting layer-wise expert modules \citep{huang2025graph} to enable finer control over structural specialization.
MoE has also been explored in multimodal graph learning. Unigraph2 \citep{he2025unigraph2} employs a basic FFN-based MoE module to align multimodal embeddings before graph propagation, but the expert networks are unaware of graph structures and thus cannot adapt to inherent relational patterns. MMGAT \citep{zhang2025moe} incorporates GNN-based experts similar to GMoE but aggregates all edge types uniformly, limiting its expressive power in graphs with rich modality-induced heterogeneity. Other attempts, such as \citep{zhou2025multimodal}, introduce the MoE module merely for processing original multimodal information, and graph learning module for regularization (e.g., Virtual Adversarial Training) rather than integrating them into the graph reasoning pipeline for downstream tasks.
Beyond performance improvements, GMoE-based approaches have also shown promise in enhancing fairness in graph learning \citep{hu2022graph,liu2023fair}.


\section{Preliminaries}
We begin by defining two types of graphs relevant to this work.
A multimodal graph is defined as \(\mathcal{G}=(\mathcal{V},\mathcal{E},\mathcal{M})\) with adjacency matrix \(\mathbf{A}\) and degree matrix \(\mathbf{D}=\deg(\mathbf{A})\), where \(\mathbf{D}_{ii}=\sum_j\mathbf{A}_{ij}\). Here, \(\mathcal{V}\) is the set of nodes, \(\mathcal{E}\subseteq\mathcal{V}\times\mathcal{V}\) is the set of edges, and \(\mathcal{M}\) represents the collection of modalities associated with the nodes. We focus on the undirected case, i.e. \(\mathbf{A}=\mathbf{A}^\top\).
Each node in the multimodal graph is associated with features derived from multiple modalities, such as text, images, audio, or other sensor data. In this work, we assume that all nodes share the same set of modalities, i.e., for all \(v \in \mathcal{V}\), the modality set equals \(\mathcal{M}\).
The multimodal node features are extracted from raw data using frozen pre-trained encoders (e.g. ViT \citep{dosovitskiy2021image} for images and Sentence Transformer \citep{reimers2019sentence} for text) and concatenated into a unified feature matrix \(\mathbf{X}\in\mathbb{R}^{n\times\sum_id_i}\), where \(n=|\mathcal{V}|\) and \(d_i\) denotes the dimensionality of modality \(i\). Each row of \(\mathbf{X}\) thus represents the concatenated multimodal feature vector for a node.
For clarity, we adopt a common multimodal setting consisting of text and visual modalities as an example in our subsequent analyses and experiments.

Another type is heterogeneous graph. It is denoted as \(\mathcal{G}^*=(\mathcal{V}^*,\mathcal{E}^*,\phi,\psi)\), where \(\mathcal{V}^*\) and \(\mathcal{E}^*\) represent the sets of nodes and edges, respectively. The node-type mapping function \(\phi:\mathcal{V}^*\rightarrow\mathcal{A}\) maps each node to its node type from the set of node types \(\mathcal{A}\), while \(\psi:\mathcal{E}^*\rightarrow\mathcal{R}\) maps each edge to its relation type from the set of relation types \(\mathcal{R}\). The adjacency matrix of a subgraph connecting type-\(\alpha\) and type-\(\beta\) nodes is denoted as \(\mathbf{A}_{\alpha,\beta}\).

Our goal is to learn low-dimensional node embeddings \(\mathbf{Z} \in \mathbb{R}^{n \times d}\) through a function \(f_\theta:\mathcal{G}, \mathbf{X} \rightarrow \mathbb{R}^{n \times d}\), parameterized by \(\theta\). This function jointly leverages the graph structure and multimodal node features to produce informative node representations.
The overall learning objective is formulated as
\begin{equation}
    \min_{\theta} \mathcal{L}(f_\theta(\mathcal{G}, \mathbf{X}), y),
\end{equation}
where \(y\) denotes optional label information under supervised or semi-supervised settings, and \(\mathcal{L}\) is a task-dependent loss function that outputs a scalar.

\section{Node Splitting and Graph Rewiring} \label{sec:nsg}

\begin{figure}
    \centering
    \includegraphics[width=0.9\linewidth]{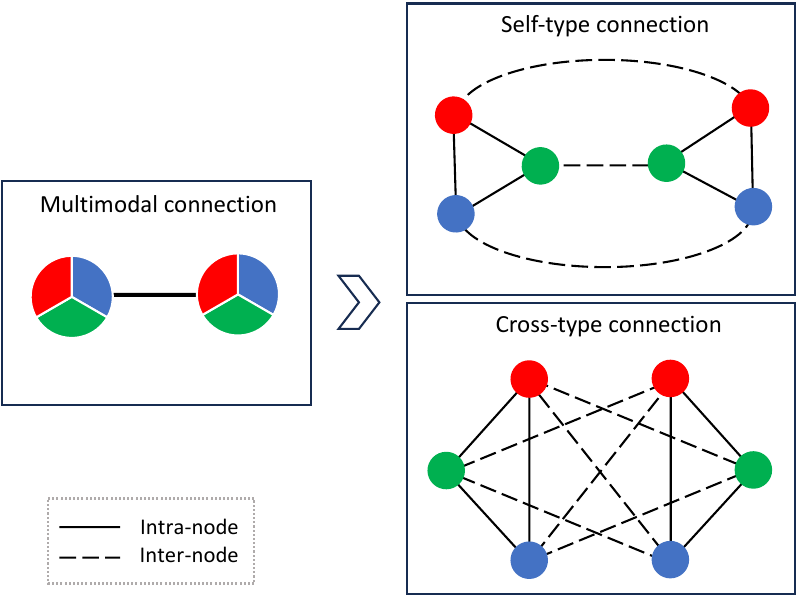}
    \caption{Illustration of node splitting and edge construction}
    \label{fig:illus}
\end{figure}

We introduce our method that transforms a homogeneous multimodal graph into a heterogeneous \underline{N}ode \underline{S}plitting \underline{G}raph (NSG) \(\mathcal{G}^*\), in which all nodes become unimodal.
This transformation enables modality-specific message passing while preserving the structural dependencies of the original graph.
The construction of NSG consists of two main stages: node splitting and graph rewiring.
Firstly, given a multimodal graph \(\mathcal{G}\), we split each node \(v\in\mathcal{V}\) into \(m=|\mathcal{M}|\) sub-nodes, denoted by \(\Omega_v=\{\langle v,1 \rangle,\dots,\langle v,m \rangle\}\), and each sub-node corresponds to a specific modality. The feature vector of each node is segmented according to modality dimensions to obtain sub-node features: \({\mathbf{X}}_t=\mathbf{X}[:,f_p(d,t):f_p(d,t+1)]\in\mathbb{R}^{n\times d_t}\), where \(f_p(d,t)=\sum_{i=1}^td_i\) defines the partition boundary for the \(t\)-th modality.
To unify feature dimensions across modalities, we introduce modality-specific linear projections: \(\mathbf{E}_t={\mathbf{X}}_t\mathbf{W}_t\), where \(\mathbf{W}_t\in\mathbb{R}^{d_t\times d}\) projects features into a shared latent space of dimension \(d\).
This process yields the complete node set of the heterogeneous graph: \(\mathcal{V}^*=\bigcup_{v\in\mathcal{V}}\Omega_v\) and the node type set \(\mathcal{T}\), which has a one-to-one correspondence with modalities in \(\mathcal{M}\).

We construct two categories of edges in \(\mathcal{G}^*\): \textit{intra-node edges} and \textit{inter-node edges}.
For a node \(v\) in the original graph, we connect every pair of its sub-nodes in \(\Omega_v\), forming a \(m\)-clique \(\mathcal{G}_v\) which has the edge set \(\mathcal{E}_v=\{(\langle v,i \rangle,\langle v,j \rangle)|1\le i<j\le m\}\).
Performing the same operation on all nodes yields the intra-node edge set \(\Phi_{\text{intra}}=\bigcup_{v\in\mathcal{V}}\mathcal{E}_v\).
These edges capture interactions among modalities of the same entity, allowing information exchange across modalities within a single node.
Inter-node edges encode the relationships between different entities. For each original edge \((u,v)\in\mathcal{E}\), we connect every sub-node of \(u\) to every sub-node of \(v\), resulting in the \emph{hybrid} edge set: \(\Phi_{u,v}^\triangle=\{(\langle u,i \rangle,\langle v,j \rangle)|1\le i\le m,1\le j\le m\}\).
Within this set, edges connecting sub-nodes of the same modality are defined as \emph{self-type} connections: \(\Phi_{u,v}^{\circ}=\{(\langle u,i \rangle,\langle v,i \rangle)|1\le i\le m\}\),
while \emph{cross-type} edges, representing cross-modal interactions, are given by \(\Phi_{u,v}^{\times}=\Phi_{u,v}^\triangle\backslash\Phi_{u,v}^{\circ}\).
In fact, these connections can be regarded as connecting two-hop neighbors based on the graph with self-type connections.
Aggregating over all node pairs yields the inter-node edge set: \(\Phi_{\text{inter}}=\bigcup_{(u,v)\in\mathcal{E}}\Phi_{u,v}\), where \(\Phi_{u,v}\) is selected from the above three options.
The final edge set of NSG is therefore \(\mathcal{E}^*=\Phi_{\text{intra}}\cup\Phi_{\text{inter}}\).
Specifically, we evaluate three configurations of NSG that differ in their inter-node edge sets, while keeping the intra-node connections consistent across all variants.
The inclusion of intra-node edges is crucial because these edges represent modality-specific correlations within the same entity, capturing the intrinsic semantic coherence between a node’s multimodal components, and simultaneously ensures structural connectivity.
Notably, if only the self-type edges \(\Phi_{u,v}^{\circ}\) are retained, NSG degenerates into \(m\) disjoint components—each corresponding to a single modality—thus recovering the structure used in prior multimodal GCNs \citep{wei2019mmgcn}.

The constructed NSG, together with the modality-projected features \(\mathbf{E}=\{\mathbf{E}_t\}\), is fed into a multi-layer HGNN to perform modality-aware message passing.
The HGNN aggregates information across both intra- and inter-node edges, producing updated embeddings for each modality type \(\mathbf{H}=\text{Enc}(\mathbf{E})=\{\mathbf{H}_t|1\le t \le m\}\).
After message propagation, we merge the split sub-nodes corresponding to each original node by concatenating their embeddings:
\begin{equation}
    \mathbf{Z}=\text{MLP}(\mathrm{Concat}_{\mathrm{col}}(\mathbf{H}_{1},\dots,\mathbf{H}_{m}))
\end{equation}
The resulting multimodal representations are then passed through a task-specific readout layer for downstream prediction tasks.

Then we analyze the computational complexity. For a general MPNN applied to the original multimodal graph, the computational complexity is \(O(L|\mathcal{E}|\sum_id_i)\), where \(L\) is the number of layers. For NSG, the number of edges satisfies \(|\mathcal{E}^*|=n\binom{m}{2}+m^2|\mathcal{E}|\). Consequently, the overall complexity becomes \(O(L|\mathcal{E}^*|d)=O(L(n+|\mathcal{E}|)d)=O(L|\mathcal{E}|d)\), which is asymptotically comparable to that of standard multimodal GNNs when \(m\) is small.
Thus, our approach maintains computational efficiency while introducing richer modality-aware graph structure.

\begin{algorithm}[t]
\caption{Approximate maximum spanning tree (MST) algorithm. Approximate-MST$(\mathbf{E}_v, c_1, c_0)$}
\label{alg:mst}
\begin{algorithmic}[1]
\Require Feature matrix $\mathbf{E}_v$ of node set $\Omega_v$, batch size $c_1$, anchor count $c_0$
\Ensure Approximate MST

\State $H \gets (V, \emptyset)$   \Comment{Edge accumulator}
\State Sample $c_0$ nodes uniformly from $V$ and set $A \gets$ these anchors 
\State $B \gets V \setminus A$

\While{$B \neq \emptyset$}

    \State Select anchor set $A_t \subseteq A$ of size $\min(c_0, |A|)$ uniformly at random
    \State Select batch $V_t \subseteq B$ of size $\min(c_1, |B|)$ uniformly at random

    \State $S \gets A_t \cup V_t$     \Comment{Local subproblem nodes}

    \State Compute pairwise cosine similarity on $S$ using $\mathbf{E}$
    
    \State $T \gets \text{MST}(S)$     \Comment{Local MST on subgraph}

    \State Add all edges of $T$ to $H$

    \State $A \gets A \cup V_t$         \Comment{Expand anchor reservoir}
    \State $B \gets B \setminus V_t$    \Comment{Remove processed nodes}

\EndWhile

\State \Return $\text{MST}(H)$   \Comment{Global pruning for final approx MST}

\end{algorithmic}
\end{algorithm}

\paragraph{Maintaining scalability when the number of modalities is large}
Although the full NSG captures fine-grained cross-modal relationships, its quadratic growth in the number of modalities (\(O(m^2)\)) can hinder scalability. To address this, we propose a sparsified NSG that preserves structural fidelity with significantly reduced edge density.
Within each node $v$, we apply the approximate MST algorithm (Algorithm \ref{alg:mst}) to obtain a tree structure.
Note that the feature dimensions are identical after the initial transformation, making the cosine similarity calculation entirely feasible.
We denote the resulting edge sets as \(\{\mathcal{E}_v^\text{spar}|v\in\mathcal{V}\}\).
For each batch, it computes an MST on at most $c_0 + c_1$ nodes, which costs \(O((c_0 + c_1)^2 \log(c_0 + c_1))\)
due to the need to compute pairwise distances and run a standard MST algorithm.
Thus, the total cost is \(O\left(\frac{m}{c_1} \cdot (c_0+c_1)^2 \log(c_0+c_1)\right)\).
Since \(c_0=O(1), c_1 \ll m\), this simplifies to \(O(mc_1 \log c_1)\).
Finally, the global pruning step computes $\text{MST}(H)$ on a sparse graph with \(O(m)\) edges, costing \(O(m \log m)\).
The overall complexity is \(O(mc_1 \log c_1 + m \log m)\), which is much faster than computing a full MST on all pairwise distances ($O(m^2)$).
The constructed tree structure preserves the connectivity and semantic relationships between nodes under the homophily assumption.
Based on this, we next address the sparsification of inter-node connections.
Directly connecting every subnode pair across two cliques not only leads to quadratic complexity, but also limiting the model's ability to exploit latent substructures.
The solution is as follows: for each node \(\langle u,i \rangle\), we connect it only to the one-hop neighbors of \(\langle v,i \rangle\) within \(\mathcal{G}_v\): $\Phi_{\langle u,i \rangle}^\times=\{(\langle u,i \rangle,\langle v,j \rangle)|$ $ (\langle v,i \rangle , \langle v,j \rangle) \in\mathcal{E}_v^\text{spar}\}$.
Taking the union over all nodes yields \(\Phi_{\text{inter}}^\times=\bigcup_{u\in\mathcal{V},i\in\mathcal{M}}\Phi_{\langle u,i \rangle}^\times\). The resulting inter-node structure has cardinality \(O(m)\), dramatically smaller than the original dense construction.
This sparsity not only reduces computational overhead but also improves generalization by discouraging over-reliance on specific edges, echoing the regularization effect observed in graph sparsification techniques such as DropEdge \citep{rong2020dropedge}.

\begin{figure}
    \centering
    \includegraphics[width=0.9\linewidth]{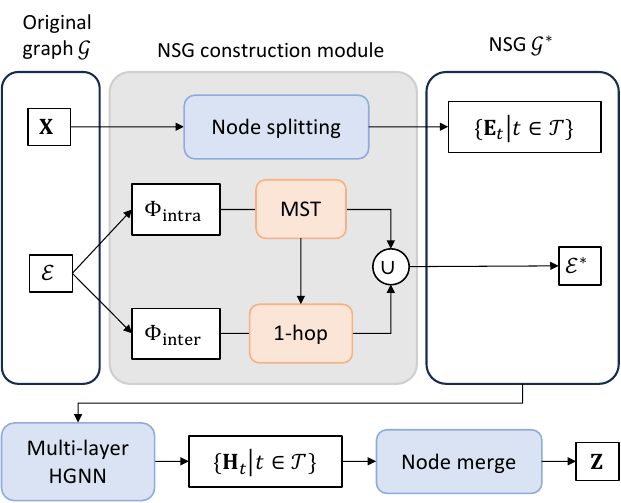}
    \caption{Procedure diagram of multimodal graph learning with NSG}
    \label{fig:nsgr}
\end{figure}

Building on these components, we present the complete NSG learning framework, illustrated in Figure~\ref{fig:nsgr}. The pipeline begins with the NSG construction module, which converts the original multimodal graph into a heterogeneous graph composed of unimodal nodes, while the two sparsification mechanisms selectively prune redundant intra- and inter-node connections to enhance efficiency. A multi-layer HGNN then performs heterogeneous message passing to aggregate modality-specific representations. Finally, a node-merging module integrates the learned subnode embeddings to obtain the representation of each original node.

\paragraph{Preliminary verification}
To preliminarily assess the effectiveness of the proposed strategy, we conduct experiments comparing three variants of NSG with the original graph (Table \ref{tab:pre}).
We adapt GraphSAGE \citep{hamilton2017inductive} for heterogeneous graphs by applying homogeneous graph convolution on each edge type to construct SAGE-h.
It can be seen that all NSG variants achieve performance improvements over the original. It is notable that the relatively slight performance improvement is limited by the expressiveness of the base model (SAGE). If we apply a stronger model, the framework's potential can be fully realized. Overall, the results confirm that decomposing nodes into unimodal sub-nodes and enabling modality-specific message passing is a more expressive and structurally sound representation paradigm.
Among the three configurations, the edge set incorporating self-type connections yields the best overall performance. This result aligns with the intuition that, for neighboring nodes in the original graph, correlations between the same modalities are generally stronger and more reliable than those across modalities. In contrast, adopting cross-type connections benefits a small number of special cases where features from different modalities are more similar. Moreover, the full inter-node edge set \(\Phi_{u,v}\) provides limited additional benefit, due to possible information blending effect.

\begin{table}[] \small
\caption{Performance comparison using different type of NSG (\%). NC stands for node classification, metric: accuracy; LP stands for link prediction, metric: MRR. Base model: GraphSAGE and extended model which applies independent GraphSAGE on each edge type; Feature encoder alignment: CLIP.}
\label{tab:pre}
\centering
\begin{tabular}{cccc}
\toprule
\multirow{2}{*}{\begin{tabular}[c]{@{}c@{}}Node/Edge   \\      set\end{tabular}} & NC & \multicolumn{2}{c}{LP} \\ \cmidrule(lr){2-2} \cmidrule(lr){3-4}
 & Ele-fashion & Sports & Cloth \\ \midrule
\((\mathcal{V},\mathcal{E})\) & 85.25 ±0.05 & 26.42 ±0.15 & 18.33 ±0.20 \\
\((\mathcal{V}^*,\Phi_{\text{intra}}\cup\Phi^{\circ})\) & 87.04 ±0.23 & 28.09 ±0.21 & 21.19 ±0.08 \\
\((\mathcal{V}^*,\Phi_{\text{intra}}\cup\Phi^\times)\) & 86.07 ±0.06 & 27.15 ±0.04 & 21.03 ±0.05 \\
\((\mathcal{V}^*,\Phi_{\text{intra}}\cup\Phi^\triangle)\) & 86.07 ±0.15 & 27.35 ±0.09 & 21.01 ±0.01
\\ \bottomrule
\end{tabular}
\end{table}



\section{Theoretical Analyses}
\subsection{A Coalesced Form of HGNNs} \label{sec:coal}
HGNNs can be viewed as a generalization of standard GNNs. To better understand the behavior of our proposed NSG architecture, we first show that a simplified form of HGNN reduces to a standard GCN operating on an expanded adjacency structure. This coalesced formulation enables a clean spectral interpretation in later sections.
We begin with a general GCN equipped with residual connections \citep{xu2018representation, chen2020simple}. For each layer \(l\) (\(l=1,\dots,L\)), the representation update is
\begin{equation} \label{eq:gcn}
\mathbf{H}^{(l)}=\sigma((\mathbf{P}+\mathbf{I})\mathbf{H}^{(l-1)}\mathbf{W}_1^{(l)}+\mathbf{H}^{(0)}\mathbf{W}_0^{(l)})
\end{equation}
where \(\sigma\) is a non-linear function, \(\mathbf{P}\in\mathbb{R}^{n\times n}\) is a propagation matrix such as the normalized adjacency \(\tilde{\mathbf{D}}^{-1/2}\tilde{\mathbf{A}}\tilde{\mathbf{D}}^{-1/2}\), and \(\mathbf{W}_0\), \(\mathbf{W}_1\in\mathbb{R}^{d\times d}\) are learnable transformations.
Following the simplifications shown effective in SGC \citep{wu2019simplifying} and GCNII \citep{chen2020simple}, we remove nonlinearities, yielding the closed-form representation
\begin{equation} \label{eq:gcnf}
    \mathbf{Z}=\mathbf{H}^{(L)}=\sum_{l=1}^L \mathbf{P}^l\mathbf{H}^{(0)}\mathbf{W}^{(l)}=\sum_{l=0}^L \mathbf{P}^l\mathbf{X}\mathbf{W}^{(l)}
\end{equation}
where \(\mathbf{W}^{(l)}\) is the sum of all weight matrices corresponding to \(\mathbf{P}^l\) in Equation \ref{eq:gcn}.
For the node classification task, applying transformation \(\mathbf{W}_c\in\mathbb{R}^{d\times C}\) and softmax function to obtain the probability of belonging to each category: \(\hat{\mathbf{Y}}=\text{softmax}(\mathbf{Z}\mathbf{W}_c)\).

HGNNs \citep{schlichtkrull2018modeling,wang2019heterogeneous,hu2020heterogeneous} typically adopt a two-level aggregation scheme:
(1) intra-relation aggregation within each meta-relation, and
(2) inter-relation aggregation across different meta-relations.
Consider a heterogeneous graph with two node types 0 and 1. Let \(\mathbf{P}_0\) and \(\mathbf{P}_1\) be propagation matrices defined on the two induced subgraphs, and \(\dot{\mathbf{P}}\) be the propagation matrix based on the cross-type interactions.
Node-level updates for type 0 at layer \(l\) are
\begin{equation}
\dot{\mathbf{H}}_0^{(l)}=\sigma(\dot{\mathbf{P}}\mathbf{H}_1^{(l-1)}\dot{\mathbf{W}}^{(l)}),
\quad
\ddot{\mathbf{H}}_0^{(l)}=\sigma(\mathbf{P}_0\mathbf{H}_0^{(l-1)}\ddot{\mathbf{W}}^{(l)}).
\end{equation}
Type-level aggregation then fuses meta-relational information:
\begin{equation}
    \mathbf{H}_0^{(l)}=\text{Aggr}(\dot{\mathbf{H}}_0^{(l)},\ddot{\mathbf{H}}_0^{(l)})\mathbf{W}_t^{(l)}
\end{equation}
To obtain a coalesced representation comparable to Eq. \eqref{eq:gcnf}, we adopt three simplifying but standard assumptions also used in prior HGNN analyses \citep{schlichtkrull2018modeling,yang2021interpretable}:
(1) remove nonlinearities;
(2) use element-wise summation as the aggregation operator;
(3) adopt a shared message encoder, i.e., \(\dot{\mathbf{W}}^{(l)}=\ddot{\mathbf{W}}^{(l)}=\mathbf{W}_h^{(l)}\).
Under these assumptions, the update becomes
\begin{equation} \label{eq:hgnn}
\begin{split}
\begin{bmatrix}
 \mathbf{H}_0^{(l)} \\
 \mathbf{H}_1^{(l)}
\end{bmatrix}
&=
\begin{bmatrix}
 \dot{\mathbf{P}}\mathbf{H}_1^{(l-1)}\mathbf{W}_h^{(l)}+\mathbf{P}_0\mathbf{H}_0^{(l-1)}\mathbf{W}_h^{(l)} \\
 \dot{\mathbf{P}}^\top\mathbf{H}_0^{(l-1)}\mathbf{W}_h^{(l)}+\mathbf{P}_1\mathbf{H}_1^{(l-1)}\mathbf{W}_h^{(l)}
\end{bmatrix}\mathbf{W}_t^{(l)}\\
&=
\underbrace{\begin{bmatrix}
 \mathbf{P}_0 & \dot{\mathbf{P}} \\
 \dot{\mathbf{P}}^\top & \mathbf{P}_1
\end{bmatrix}}_{\text{propagation matrix}}
\begin{bmatrix}
 \mathbf{H}_0^{(l-1)} \\
 \mathbf{H}_1^{(l-1)}
\end{bmatrix}
\underbrace{\mathbf{W}_h^{(l)}
\mathbf{W}_t^{(l)}}_{\text{weight matrix}}
\end{split}
\end{equation}
Crucially, the first term acts exactly as a propagation matrix on an expanded homogeneous graph.
When \(\mathbf{P}_0\), \(\mathbf{P}_1\) and \(\dot{\mathbf{P}}\) are taken as adjacency matrices of the intra- and inter-type subgraphs, it becomes identical to the adjacency structure of NSG.
Thus, a simplified HGNN layer is mathematically equivalent to a standard GCN layer on NSG, with appropriate block-structured propagation.
Unrolling Eq. \eqref{eq:hgnn} over $L$ layers and regrouping coefficients yields a closed form analogous to Eq. \eqref{eq:gcnf}, establishing the equivalence.

In NSG, each original multimodal node is split into unimodal nodes whose final embeddings \(\mathbf{H}_0^{(L)}\) and \(\mathbf{H}_1^{(L)}\) share dimensionality.
The node-merge operation is therefore a simple concatenation followed by a linear transformation:
\begin{equation}
    \mathbf{Z}=
\left[
\begin{array}{c|c}
\mathbf{H}_0^{(L)} & \mathbf{H}_1^{(L)} \\
\end{array}
\right].
\end{equation}
This completes the reduction of the NSG learning procedure to a coalesced GCN defined over a block-structured propagation matrix, laying the groundwork for the spectral analysis that follows.

\subsection{Spectral Analyses of NSG}
Prior spectral analyses of HGNNs typically treat each node type as an independent signal domain, constructing separate Laplacians for each type-specific subgraph and then combining their outputs \citep{yu2022multiplex,butler2023convolutional,he2024spectral}.
Although intuitive, this approach implicitly assumes that signals belonging to different node types evolve in disconnected spectral spaces, preventing cross-type coupling in the frequency domain.
Such decoupling obscures the true joint propagation behavior in heterogeneous graphs, especially when message passing occurs across node types.
Building on the coalesced HGNN formulation derived in Section \ref{sec:coal}, we analyze the proposed model directly in the joint spectral space of the entire heterogeneous graph. This unified view enables us to characterize how multimodal signals—represented as node-type–specific components—interact through both self-type and cross-type propagation.

To make the analysis concrete, we consider a symmetric setting in which each modality-specific node set has $n$ nodes and identical within-modality connectivity.
Let \(\mathbf{A}\) denote the adjacency among nodes of the same modality (self-type edges), and \(\mathbf{B}\) denote adjacency between nodes of different modalities (cross-type edges, including intra-node links).
For NSG, \(\mathbf{B}\in \{\mathbf{I}, \mathbf{I}+\mathbf{A}\}\), corresponding to the three edge sets \(\Phi_{\text{intra}}\) and \(\Phi_{\text{intra}}\cup\Phi^\times\).
The full adjacency becomes:
\begin{equation}
\text{Adj}=
\begin{bmatrix}
    \mathbf{A} & \mathbf{B} \\
    \mathbf{B} & \mathbf{A} \\
\end{bmatrix},
\end{equation}
and after applying any degree-based or attention-based normalization, we obtain:
\begin{equation}
\widehat{\text{Adj}}=
\begin{bmatrix}
    \hat{\mathbf{A}} & \hat{\mathbf{B}} \\
    \hat{\mathbf{B}} & \hat{\mathbf{A}} \\
\end{bmatrix}.
\end{equation}
The corresponding (normalized) Laplacian is:
\begin{equation}
    \mathbf{L}=\mathbf{I}_{2n}-\widehat{\text{Adj}}=
\begin{bmatrix}
    \mathbf{I}_n-\hat{\mathbf{A}} & -\hat{\mathbf{B}} \\
    -\hat{\mathbf{B}} & \mathbf{I}_n-\hat{\mathbf{A}} \\
\end{bmatrix}
\end{equation}
This matrix admits a closed-form block diagonalization due to its symmetric two-block structure. Using an orthogonal matrix \(\mathbf{Q}\), we obtain the transformed Laplacian:
\begin{equation} \label{eq:diag}
\mathbf{Q}^\top\mathbf{L}\mathbf{Q}=
\begin{bmatrix}
    \mathbf{I}_n-\hat{\mathbf{A}}-\hat{\mathbf{B}} & \mathbf{0} \\
    \mathbf{0} & \mathbf{I}_n-\hat{\mathbf{A}}+\hat{\mathbf{B}} \\
\end{bmatrix},
\mathbf{Q}=\frac{1}{\sqrt{2}}
    \begin{bmatrix}
        \mathbf{I}_n & \mathbf{I}_n \\
        \mathbf{I}_n & -\mathbf{I}_n \\
    \end{bmatrix}
\end{equation}
Let \(\mathbf{U}_1\) diagonalize \(\mathbf{I}_n-\hat{\mathbf{A}}-\hat{\mathbf{B}}\) and \(\mathbf{U}_2\) diagonalize \(\mathbf{I}_n-\hat{\mathbf{A}}+\hat{\mathbf{B}}\), then
\begin{equation}
\begin{split}
\mathbf{Q}^\top\mathbf{L}\mathbf{Q}&=
\begin{bmatrix}
    \mathbf{U}_1\mathbf{\Lambda}_1\mathbf{U}_1^\top & \mathbf{0} \\
    \mathbf{0} & \mathbf{U}_2\mathbf{\Lambda}_2\mathbf{U}_2^\top \\
\end{bmatrix}\\
&=
\begin{bmatrix}
    \mathbf{U}_1 & \mathbf{0} \\
    \mathbf{0} & \mathbf{U}_2 \\
\end{bmatrix}
\begin{bmatrix}
    \mathbf{\Lambda}_1 & \mathbf{0} \\
    \mathbf{0} & \mathbf{\Lambda}_2 \\
\end{bmatrix}
\begin{bmatrix}
    \mathbf{U}_1 & \mathbf{0} \\
    \mathbf{0} & \mathbf{U}_2 \\
\end{bmatrix}^\top.
\end{split}
\end{equation}
Thus, the eigenvectors of \(\mathbf{L}\) take the form:
\begin{equation}
    \mathbf{u}\in\mathcal{F}_1 \Leftrightarrow \mathbf{u} = \tfrac{1}{\sqrt{2}}\begin{bmatrix}\mathbf{v}\\ \mathbf{v}\end{bmatrix},\qquad
\mathbf{u}\in\mathcal{F}_2 \Leftrightarrow \mathbf{u} = \tfrac{1}{\sqrt{2}}\begin{bmatrix}\mathbf{v}\\ -\mathbf{v}\end{bmatrix},
\end{equation}
where \(\mathbf{v}\) is an eigenvector of the corresponding block in \eqref{eq:diag}.
Eigenspace \(\mathcal{F}_1\) captures frequency components shared across modalities, while \(\mathcal{F}_2\) captures discrepancy components encoding differences between modalities.
The full eigenbasis is
\begin{equation}
    \mathbf{U}=\mathbf{Q}\, \mathrm{diag}(\mathbf{U}_1,\mathbf{U}_2),
\end{equation}
and defines the Fourier transform \(\mathbf{x}_\mathcal{F}=\mathbf{U}^\top\mathbf{x}\).

We now analyze the spectral behavior of the simplified HGNN propagation in Section \ref{sec:coal}. Let: \(\mathbf{P}_0=\mathbf{P}_1=\alpha\mathbf{I}+\hat{\mathbf{A}}\), \(\dot{\mathbf{P}}=\beta\mathbf{I}+\hat{\mathbf{B}}\), where \(\alpha\) models adaptive self-loops (zeroth-order Chebyshev term), and \(\beta\) scales intra-node information flow.
Then the graph filter is formulated as
\begin{equation}
\mathbf{U}{\mathbf{G}}\mathbf{U}^\top=
    \begin{bmatrix}
\alpha\mathbf{I}+\hat{\mathbf{A}} & \beta\mathbf{I}+\hat{\mathbf{B}} \\
\beta\mathbf{I}+\hat{\mathbf{B}} & \alpha\mathbf{I}+\hat{\mathbf{A}} \\
    \end{bmatrix},
\end{equation}
Applying the same basis change as in \eqref{eq:diag} yields:
\begin{equation}
    {\mathbf{G}}=
    \begin{bmatrix}
(\alpha+\beta+1)\mathbf{I}-\mathbf{\Lambda}_1 & \mathbf{0} \\
\mathbf{0} & (\alpha-\beta+1)\mathbf{I}-\mathbf{\Lambda}_2  \\
    \end{bmatrix}
\end{equation}
Hence, the frequency response function in the two subspaces is defined based on a tuple of eigenvalue and eigenvector:
\begin{equation}
    h(\lambda;\mathbf{u})=
\begin{cases}
 \alpha+\beta+1-\lambda & \text{ if } \mathbf{u}\in\mathcal{F}_1 \\
 \alpha-\beta+1-\lambda & \text{ if } \mathbf{u}\in\mathcal{F}_2
\end{cases}.
\end{equation}
This result shows that the propagation acts as a joint low-pass filter on the heterogeneous graph, yet with distinct filtering profiles in the two modality-induced spectral subspaces:
The \(\mathcal{F}_1\) subspace captures features shared across modalities; the gain increases with \(\beta\), strengthening alignment.
The \(\mathcal{F}_2\) subspace captures modality differences; the gain decreases with \(\beta\), suppressing modality-specific noise.
Thus, \(\beta\) explicitly tunes the balance between cross-modality smoothing and the preservation of modality-specific distinctions.
Meanwhile, \(\alpha\) uniformly shifts both responses without altering the difference between subspaces.
Finally, we note that renormalization tricks such as adding pre-normalized self-loops \citep{kipf2017semi} are not directly applicable in heterogeneous graphs, where degrees mix contributions from different meta-relations. Treating self-loops as an independent relation and normalizing per edge type preserves the semantic structure of modalities and ensures stable training.



\subsection{Comparing With General GNN Models}
In multimodal graph learning, the proposed NSG architecture induces a more structured hypothesis class than general GNNs. This structural restriction plays a beneficial role in controlling model complexity and, consequently, improving generalization. We formalize this intuition by comparing the information-theoretic generalization bounds of the two model families.

We measure the generalization error of a learning model \(f\) on an underlying data distribution \(\mathcal{D}\) as
\begin{equation}
    R(f)=\mathbb{E}_{(x,y)\sim\mathcal{D}}[L(f,x,y)]
\end{equation}
and define \(\hat{R}(f)\) as the empirical training loss over a sample set \(S=\{s_1,\dots,s_n\}\).

We use the mutual-information–based bound from \citep{xu2017information}:
\begin{lemma}[Information-theoretic generalization bound] \label{lem:info}
For a model with parameters \(W\) trained on sample set \(S\),
\[
|\hat{R}(f)-R(f)|\le\sqrt{\frac{2\sigma^2}{n}I(S;W)}
\]
where \(\sigma^2\) is a sub-Gaussian constant of the loss function.
\end{lemma}
Thus, the generalization ability is directly linked to the mutual information \(I(S;W)\), which quantifies how sensitively the learned parameters depend on the training samples.
Let \(\Theta_\text{GNN}\) denote the parameter space of a general GNN, and \(\Theta_\text{NSG}\) denote the parameter space realizable under the NSG architecture.
As shown in Section \ref{sec:coal}, NSG enforces a structured parameterization that prevents arbitrary mixing of the two modalities.
Concretely, while a general GNN may include transformations of the form
\[
\begin{bmatrix}
    \mathbf{E}_1 & \mathbf{E}_2 
\end{bmatrix}\mathbf{W}, \quad \mathbf{W}\in\mathbb{R}^{2d\times 2d},
\]
NSG only allows block-diagonal transformations \(\mathbf{W}=\begin{bmatrix}
    \mathbf{W}_1 & \mathbf{0} \\
    \mathbf{0} & \mathbf{W}_2 
\end{bmatrix}\),
so that no cross-modality mixing can occur inside each HGNN layer. This restriction is fundamental: it ensures that all cross-modality interactions must pass through explicitly designed structural connections (e.g., intra-node edges), rather than uncontrolled parameter entanglement.
Formally, the parameter spaces satisfy:
\(\Theta_\text{NSG}\subseteq\Theta_\text{GNN}\).
Therefore, for every NSG parameter setting \(W_{\text{NSG}}\in \Theta_\text{NSG}\), there exists a (not necessarily unique) \(W_{\text{GNN}}\in \Theta_\text{GNN}\) and a measurable transformation \(T:\Theta_\text{GNN}\to\Theta_\text{NSG}\) such that \(W^{\text{NSG}}=T(W_{\text{GNN}})\).
This induces a Markov chain \(S\to W_{\text{GNN}} \to W_{\text{NSG}}\).
From the Markov chain and the data processing inequality, we obtain:
\begin{equation}
    I(S;W_{\text{NSG}})=I(S;T(W_{\text{GNN}}))\le I(S;W_{\text{GNN}})
\end{equation}
Thus, the NSG architecture, by restricting the parameter space, reduces the mutual information between the sample set and the learned parameters. This effect directly yields a tighter generalization guarantee.
Combining with Lemma \ref{lem:info}, we obtain that the NSG model has a better (i.e., lower) generalization bound than arbitrary GNNs.

\begin{figure*}[t]
    \centering
    \includegraphics[width=\linewidth]{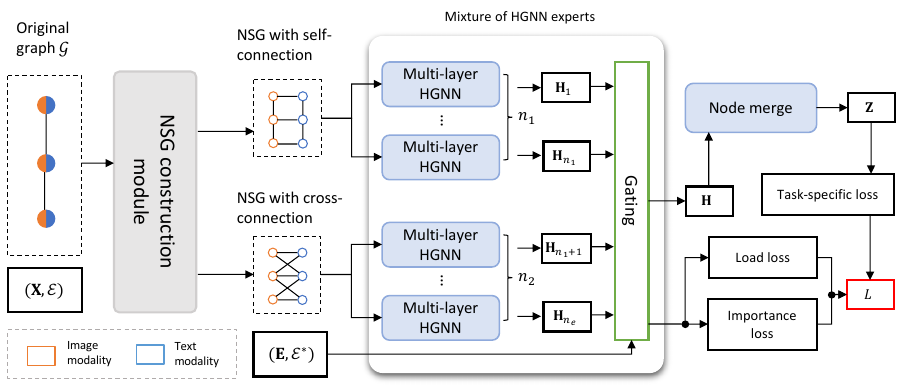}
    \caption{Overall framework of the GMoE-extended model}
    \label{fig:fw}
\end{figure*}

\section{Cost-effective Mixture of HGNN Experts}
As discussed in Section~\ref{sec:nsg}, the inter-node edge set in the NSG can be constructed in several ways. Each construction captures different relational structures and thus exhibits distinct advantages depending on the underlying data characteristics. A natural question arises: \emph{can we integrate the strengths of multiple NSG variants within a unified learning framework to further enhance performance?}
To this end, we propose NSG-MoE, a cost-effective Mixture of HGNN Experts model. This design draws inspiration from the mixture-of-experts (MoE) paradigm, which has proven highly effective in improving representational diversity of the model.
Similar to GMoE \citep{wang2023graph}, which employs distinct experts to handle neighbors from different hop distances, our framework introduces a collection of HGNN-based experts, each specialized in processing a specific variant of NSG. Unlike GMoE focusing on message depth, NSG-MoE explicitly models modality-dependent relational diversity, learning from selectively constructed one- or two-hop multimodal structures.

\paragraph{Dual-Branch Expert Design}
We focus on two primary NSG variants: one containing self-type inter-node connections and the other containing cross-type connections. This dual-branch design achieves a balance between model expressiveness and computational cost by discarding more time-consuming and less effective hybrid edge set.
Corresponding to these two edge types, we define two groups of experts.
Suppose there are \(n_1\) and \(n_2\) experts handling self-type and cross-type structures respectively, with total \(n_e=n_1+n_2\). The experts are indexed by \(e=1,\dots,n_e\). The first \(n_1\) HGNN encoders take \((\{\mathbf{E}_t\},\Phi_{\text{intra}}\cup\Phi_{\text{inter}}^\circ)\) as input, while the remaining \(n_2\) encoders take \((\{\mathbf{E}_t\},\Phi_{\text{intra}}\cup\Phi_{\text{inter}}^\times)\). Each expert outputs an embedding \(\mathbf{H}_e\), and the final node representation for sub-node \(\langle u,t \rangle\) is computed as:
\begin{equation}
    \mathbf{H}_{\langle u,t \rangle}^{\text{moe}}=\sum_{e=1}^{n_e} [G(\mathbf{E}_{\langle u,t \rangle})]_e(\mathbf{H}_e)_{\langle u,t \rangle}
\end{equation}
where \(\mathbf{E}_{\langle u,t \rangle}\) denotes the feature vector of node \(u\) in modality \(t\), and \(G(\cdot)\) is the gating function that dynamically assigns routing probabilities over experts.
In practice, all experts share the same HGNN architecture and hyperparameters, but each maintains independent parameters. Consequently, even within the same group, experts can specialize in different sub-distributions of input data, thereby capturing diverse modality-specific patterns.
\paragraph{Noisy Gating Mechanism}
The gating function \(G:\mathbb{R}^{d}\rightarrow\mathbb{R}^{n_1+n_2}\) determines which experts are most suitable for processing a given input. It introduces controlled stochasticity to promote specialization and balanced utilization across experts. Formally, for any input \(x\in\mathbb{R}^{d}\),
\begin{equation}
    G(x)=\mathrm{Softmax}(\mathrm{TopK}(S(x),k_e)),
\end{equation}
where
\begin{equation}
    S(x)=x W_g + \epsilon \cdot \mathrm{Softplus}(x W_n).
\end{equation}
Here, \(\epsilon\in\mathbb{R}^{n_e}\) is a vector of i.i.d. standard normal variables, \(W_g,W_n\in\mathbb{R}^{d\times n_e}\) are learnable matrices initialized to zero to maintain equal expert loads at the start of training, and \(\mathrm{TopK}\) retains only the \(k_e\) largest elements while setting others to \(-\infty\).


\paragraph{Expert Balancing Objectives}
In order to promote balanced expert utilization, we introduce two auxiliary regularization losses.
First, we constrain the importance imbalance—the difference in total routing probabilities assigned to different experts—using:
\begin{equation}
    L_{\text{importance}}=[CV(\sum_{\langle u,t \rangle\in\mathcal{V}^*}G(\mathbf{E}_{\langle u,t \rangle}))]^2,
\end{equation}
where \(CV(\cdot)\) denotes the coefficient of variation. A lower \(L_{\text{importance}}\) indicates that experts contribute more evenly to the final output.
Second, to alleviate the load imbalance problem \citep{shazeer2017}, we introduce:
\begin{equation}
    L_{\text{load}}=[CV([\sum_{\langle u,t \rangle\in\mathcal{V}^*} P(\mathbf{E}_{\langle u,t \rangle},e)])]^2,
\end{equation}
where \(P(x,e)\) represents the probability that expert is selected when \(\epsilon_e\) is resampled for input \(x\):
\begin{equation}
    P(x,e)=\Phi\left ( \frac{[x W_g]_e - \mathrm{topk}(S(x),e,k_e)}{\mathrm{Softplus}([x W_n]_e)} \right )
\end{equation}
and \(\Phi(\cdot)\) is the CDF of the standard normal distribution. Here $\mathrm{topk}(\cdot,e,k)$ returns the $k$-th largest element excluding the \(e\)-th element.
While \(L_{\text{importance}}\) regulates the overall output distribution of the gating network, \(L_{\text{load}}\) focuses on the frequency of expert activation, together ensuring that all experts receive sufficient gradient signals during training.
The overall training objective combines the task-specific loss \(L_{\text{task}}\) with these two balancing terms:
\begin{equation}
L=L_{\text{task}}+\lambda(L_{\text{importance}}+L_{\text{load}}),
\end{equation}
where \(\lambda\) controls the regularization strength.

The overall architecture is illustrated in Figure~\ref{fig:fw}.
Compared with the basic model, NSG-MoE simultaneously leverages two complementary NSG structures and replaces the single HGNN encoder with a mixture of HGNN experts. This enables the model to capture complex substructures formed by modality-specific and cross-modality interactions across nodes.
In large-scale graphs, although MoE models can theoretically lead to sparse activation, the large number of nodes ensures that nearly all experts are activated in each forward pass.
Nevertheless, given the relative sparsity of the original graph’s edges, the degree of each node will not be too large. Therefore, only a modest number of experts is needed to capture fine-grained multimodal semantics effectively.
The computational complexity of the full model scales as \(O(n_eLdm|\mathcal{E}|)\) which remains cost-effective when \(n_e\) is small.
Importantly, NSG-MoE is highly modular, as each expert can be replaced by any heterogeneous GNN architecture. This flexibility allows NSG-MoE to serve as a universal framework for multimodal graph learning, easily adaptable to node-level or edge-level prediction tasks through the selection of an appropriate task-specific loss.
In addition, we emphasize that the NSG method is naturally applicable to MoE-style extensions, as it generates two forms of structure, which is absent in other methods.

\begin{table}[t] \small
\centering
\caption{Dataset Statistics.}
\label{tab:data}
\begin{tabular}{ccccc}
\toprule
Dataset & \#Nodes & \#Edges & \#Classes & \#Modality \\ \midrule
Ele-fashion & 97,766 & 199,602 & 12 & 2 \\
Movies & 16,672 & 218,390 & 20 & 2 \\
Toys & 20,695 & 126,886 & 18 & 2 \\
Grocery & 17,074 & 171,340 & 20 & 2 \\
Reddit-S & 15,894 & 566,160 & 20 & 2 \\
Reddit-M & 99,638 & 1,167,188 & 50 & 2 \\
ABIDE & 871 & - & 2 & 4 \\ \midrule
Amazon-Sports & 50,250 & 356,202 & - & 2 \\
Amazon-Cloth & 125,839 & 951,271 & - & 2
\\ \bottomrule
\end{tabular}
\end{table}

\begin{table*}[]
\caption{results node classification. For all metrics, the larger the better. The results are reported in percentage, the same below. The best results are bolded, and the second-best are underlined. The performance improvements are significant under t-test with \(p\)-value \(\le 0.05\).}
\label{tab:nc}
\begin{subtable}{0.7\textwidth}
\centering
\caption{MAGB. The underlined results are the second-best using \emph{Llama+CLIP} feature.}
\label{tab:magb}
\begin{tabular}{cccccc}
\toprule
Model & Movies & Toys & Grocery & Reddit-S & Reddit-M \\ \midrule
\multicolumn{6}{c}{\cellcolor[HTML]{E8E8E8}Qwen-VL-7B} \\
GCN & 51.12 ±0.33 & 78.70 ±0.07 & 81.56 ±0.20 & 93.98 ±0.21 & 79.22 ±0.13 \\
SAGE & 50.83 ±0.14 & 78.47 ±0.33 & 81.73 ±0.51 & 94.28 ±0.08 & 79.48 ±0.18 \\
GAT & 52.43 ±0.25 & 78.51 ±0.22 & 81.31 ±0.04 & 94.94 ±0.08 & 79.87 ±0.38 \\ \midrule
\multicolumn{6}{c}{\cellcolor[HTML]{E8E8E8}Llama-11B} \\
GCN & 49.92 ±0.57 & 80.72 ±0.16 & 85.31 ±0.19 & 95.44 ±0.12 & 83.62 ±0.08 \\
SAGE & 49.76 ±0.18 & 80.67 ±0.47 & 85.13 ±0.17 & 95.60 ±0.09 & 84.22 ±0.10 \\
GAT & 50.26 ±0.22 & 80.00 ±0.68 & 84.72 ±0.25 & 95.87 ±0.26 & 84.71 ±0.12 \\ \midrule
\multicolumn{6}{c}{\cellcolor[HTML]{E8E8E8}Llama-8B+CLIP} \\
GCN & 47.68 ±0.73 & 78.96 ±0.15 & 84.83 ±0.14 & 90.36 ±0.52 & 62.32 ±0.34 \\
SAGE & 46.41 ±0.33 & 79.27 ±0.27 & 85.12 ±0.18 & 90.24 ±0.13 & 62.75 ±0.29 \\
GAT & 47.42 ±0.67 & 78.58 ±0.26 & 84.67 ±0.19 & 92.47 ±0.22 & 69.27 ±0.30 \\
MMGCN & {\ul 50.72 +0.32} & 80.05 +0.32 & 85.30 +0.20 & 92.43 +0.10 & {\ul 79.24 +0.33} \\
MMGAT & 49.32 +0.18 & {\ul 80.59 +0.17} & {\ul 85.60 +0.23} & {\ul 94.32 +0.07} & 79.11 +0.13 \\
Unigraph2 & 48.48 +0.23 & 79.49 +0.16 & 85.24 +0.10 & 93.50 +0.38 & 70.19 +0.38 \\
NSG-MoE & \textbf{53.98 ±0.22} & \textbf{80.92 ±0.37} & \textbf{86.28 ±0.16} & \textbf{95.93 ±0.15} & \textbf{85.03 ±0.28}
\\ \bottomrule
\end{tabular}
\end{subtable}
\hfill
\begin{subtable}{0.25\textwidth}
\centering
\caption{MM-Graph}
\begin{tabular}{cc}
\toprule
Model & Ele-fashion \\ \midrule
\multicolumn{2}{c}{\cellcolor[HTML]{E8E8E8}CLIP} \\
GCN & 85.18 ±0.09 \\
SAGE & 85.25 ±0.05 \\
GAT & 85.05 ±0.12 \\
MMGCN & 86.08 ±0.12 \\
MMGAT & {\ul 86.18 ±0.21} \\
Unigraph2 & 86.01 ±0.19 \\
NSG-MoE & \textbf{87.40 ±0.24} \\ \midrule
\multicolumn{2}{c}{\cellcolor[HTML]{E8E8E8}ImageBind} \\
GCN & 85.94 ±0.11 \\
SAGE & 85.74 ±0.14 \\
GAT & 86.29 ±0.19 \\
MMGCN & 87.21 ±0.15 \\
MMGAT & {\ul 87.86 ±0.17} \\
Unigraph2 & 86.75 ±0.13 \\
NSG-MoE & \textbf{88.42 ±0.11}
\\ \bottomrule
\end{tabular}
\end{subtable}
\end{table*}

\begin{table*}[t] \small
\caption{Results on link prediction task.}
\label{tab:lp}
\begin{tabular}{ccccccccc}
\toprule
 & \multicolumn{4}{c}{Amazon-Sports} & \multicolumn{4}{c}{Amazon-Cloth} \\ \cmidrule(lr){2-5} \cmidrule(lr){6-9}
\multirow{-2}{*}{Model} & H@1 & H@3 & H@10 & MRR & H@1 & H@3 & H@10 & MRR \\ \midrule
\multicolumn{9}{c}{\cellcolor[HTML]{E8E8E8}CLIP as feature   encoder alignment} \\
GCN & 14.60 ±0.11 & 30.16 ±0.32 & 56.03 ±0.50 & 27.59 ±0.20 & 11.11 ±0.07 & {\ul 22.18 ±0.13} & {\ul 42.10 ±0.33} & {\ul 21.29 ±0.13} \\
SAGE & 13.32 ±0.18 & 27.69 ±0.26 & 55.08 ±0.34 & 26.42 ±0.15 & 9.54 ±0.12 & 19.42 ±0.28 & 38.59 ±0.35 & 18.33 ±0.20 \\
GAT & 14.37 ±0.06 & 29.76 ±0.19 & 58.47 ±0.08 & 27.77 ±0.09 & 9.84 ±0.05 & 20.05 ±0.07 & 39.21 ±0.02 & 19.54 ±0.04 \\
MMGCN & 14.42 ±0.09 & 29.92 ±0.12 & 55.97 ±0.35 & 27.51 ±0.19 & 10.86 ±0.41 & 21.71 ±0.14 & 40.97 ±0.23 & 21.24 ±0.18 \\
MMGAT & {\ul 14.88 ±0.12} & {\ul 30.43 ±0.13} & {\ul 59.02 ±0.36} & {\ul 27.93 ±0.23} & {\ul 11.21 ±0.27} & 21.75 ±0.19 & 41.87 ±0.32 & 20.62 ±0.44 \\
Unigraph2 & 14.62 ±0.29 & 29.67 ±0.13 & 58.57 ±0.38 & 27.75 ±0.32 & 10.12 ±0.18 & 20.17 ±0.28 & 39.49 ±0.15 & 19.73 ±0.41 \\
NSG-MoE & \textbf{16.02 ±0.18} & \textbf{33.11 ±0.19} & \textbf{61.27 ±0.18} & \textbf{29.97 ±0.18} & \textbf{11.95 ±0.07} & \textbf{23.82 ±0.20} & \textbf{44.85 ±0.18} & \textbf{22.64 ±0.13} \\ \midrule
\multicolumn{9}{c}{\cellcolor[HTML]{E8E8E8}ImageBind as feature   encoder alignment} \\
GCN & 18.15 ±0.07 & {\ul 36.27 ±0.06} & 64.03 ±0.10 & 32.44 ±0.07 & 11.53 ±0.08 & 23.78 ±0.15 & 45.14 ±0.20 & 22.59 ±0.12 \\
SAGE & 14.30 ±0.05 & 29.21 ±0.13 & 56.17 ±0.21 & 27.21 ±0.10 & 10.33 ±0.07 & 20.62 ±0.20 & 39.66 ±0.36 & 20.05 ±0.14 \\
GAT & 16.74 ±0.08 & 34.01 ±0.14 & 63.43 ±0.11 & 30.95 ±0.03 & 11.68 ±0.03 & 23.11 ±0.03 & 43.84 ±0.15 & 22.15 ±0.02 \\
MMGCN & 18.23 ±0.09 & 35.69 ±0.15 & 62.17 ±0.23 & 31.86 ±0.08 & 11.51 ±0.22 & 23.68 ±0.20 & 45.36 ±0.06 & 22.26 ±0.07 \\
MMGAT & {\ul 18.67 ±0.24} & 36.03 ±0.18 & {\ul 64.96 ±0.32} & {\ul 33.16 ±0.16} & {\ul 12.18 ±0.07} & {\ul 23.80 ±0.04} & {\ul 45.90 ±0.18} & {\ul 22.94 ±0.10} \\
Unigraph2 & 17.34 ±0.10 & 35.85 ±0.33 & 63.37 ±0.09 & 31.27 ±0.14 & 11.77 ±0.03 & 23.62 ±0.06 & 44.21 ±0.05 & 22.50 ±0.11 \\
NSG-MoE & \textbf{19.39 ±0.13} & \textbf{38.71 ±0.17} & \textbf{68.41 ±0.08} & \textbf{34.39 ±0.13} & \textbf{12.82 ±0.03} & \textbf{25.52 ±0.04} & \textbf{47.87 ±0.06} & \textbf{24.10 ±0.02}
\\ \bottomrule
\end{tabular}
\end{table*}

\section{Experiments}
\subsection{Datasets}
To evaluate the effectiveness and generalization capability of our proposed NSG-MoE model, we conduct extensive experiments on two multimodal graph learning benchmarks: MM-Graph \citep{zhu2025mosaic}, MAGB \citep{yan2025graph}, and one independent dataset ABIDE \citep{zheng2022multi}.
For the node classification task, we select one dataset from MM-Graph (Ele-Fashion), five datasets from MAGB (Movies, Toys, Grocery, Reddit-M, and Reddit-S), and ABIDE. For the link prediction task, we use two datasets (Amazon-Sports and Amazon-Cloth) from MM-Graph.
The original features in MM-Graph benchmark are encoded by CLIP \citep{radford2021learning} or Imagebind \citep{girdhar2023imagebind}, and MAGB employs multimodal large language models (MLLMs) as the feature encoders.
The MMGraph and MAGB datasets each contain two modalities (visual and text), while ABIDE involves four modalities.
Since ABIDE does not provide predefined graph edges, following the protocol in \citep{zheng2022multi}, we construct the adjacency matrix using cosine similarity among node features and connect each node to its $k=5$ nearest neighbors via a kNN procedure.
Dataset statistics are shown in Table \ref{tab:data}.

\subsection{Baselines}
Given the emerging nature of multimodal graph learning, the number of directly comparable baselines remains limited. We select three representative and competitive methods that cover a range of multimodal GNN paradigms:
MMGCN \citep{wei2019mmgcn}, originally proposed for recommendation tasks. Although designed for item-user interactions, its learned node embeddings can be readily adapted for general downstream graph tasks such as classification and link prediction.  
Unigraph2 \citep{he2025unigraph2}, which includes a MoE module where each expert is implemented as a MLP. Since it incorporates GraphMAE \citep{hou2022graphmae} for pretraining, which provides additional performance gains unrelated to the model architecture, we remove this component to ensure a fair comparison of all models.
MMGAT \citep{zhang2025moe} is originally designed to handle three modalities, where features of two modalities contribute to another, thus the modalities are asymmetric. We extend their approach by allowing each modality to contribute to the others.
We also incorporate three basic graph learning models GCN \citep{kipf2017semi}, GraphSAGE \citep{hamilton2017inductive}, GAT \citep{velivckovic2018graph} as baselines.

\subsection{Implementation details and hyperparameters}
For all HGNN models, we tune the number of message-passing layers \{1, 2, 3, 5\} to mitigate over-smoothing effects. Hidden dimensionality is searched within \{32, 64, 128\} and the number of attention heads—when applicable—is selected from \{1, 2, 4, 8\}.
Residual connections \citep{he2016deep} and GraphNorm layers \citep{cai2021graphnorm} are incorporated after each layer to improve numerical stability and gradient propagation.
Optimization is performed using the Adam optimizer \citep{kingma2015adam} with a learning rate selected from \{5e-5, 1e-4, 2e-4\} and weight decay from \{0.0, 1e-4\}.
Unless otherwise stated, we set the number of experts in our MoE module as \(n_1=n_2=2\) with \(k=2\) active experts.
For other models that include the MoE module, we set the total number of experts to 4 and the number of activated experts to 2.
In the main experiments, we implemented the HGNN module as HGT \citep{hu2020heterogeneous}.
We conducted all experiments on a server equipped with 256GB RAM and a single NVIDIA RTX 6000 Ada (48GB) GPU. The software environment is Ubuntu 22.04.4 LTS.

For node classification, we consider a semi-supervised setting where only a subset of nodes \(\mathcal{V}_L\subset\mathcal{V}\) are labeled. Let \(Y_L \in \{0,1\}^{|V_L| \times C}\) denote the one-hot label matrix, where \(C\) is the number of node classes. The task loss is defined as the standard cross-entropy loss between predictions \(\mathbf{Z}\) and ground-truth labels \(Y_L\).
For link prediction, we follow the setup in \citep{kipf2016variational}, using an inner-product decoder followed by a sigmoid activation to estimate edge probabilities:
\begin{equation}
\hat{A}_{ij}=\sigma(\mathbf{Z}_i^\top\mathbf{Z}_j)
\end{equation}
The binary cross-entropy loss is then computed as:
\begin{equation}
    L_{lp}=-\sum_{(i,j)\in\mathcal{E}_{\text{train}}}[A_{ij}\log\hat{A}_{ij}+(1-A_{ij})\log(1-\hat{A}_{ij})]
\end{equation}
The entire model, including the NSG-MoE module and its gating function, is trained end-to-end via backpropagation to minimize the joint objective \(L\).
\footnote{Our code is available at github.com/zyh23-chn/NSG-MoE.}

\subsection{Main Results}
We evaluate node classification using accuracy, and link prediction using Hits@K and mean reciprocal rank (MRR).
Table \ref{tab:nc} reports the mean and standard deviation over three runs for node classification. Several consistent observations emerge from these results.
(1) Across both MMGraph and MAGB, the proposed NSG-MoE achieves the strongest performance when compared under the same feature encoder. On the MAGB datasets, NSG-MoE exceeds the second-best model by an average relative improvement of over 3.33\%, and by 1.03\% on MMGraph. These gains confirm that decoupling message passing across modalities—central to NSG’s design—reduces cross-modal interference and allows the model to capture modality-specific structural dependencies more effectively.
(2) A recurring pattern across datasets is that modality separation is essential for multimodal graph learning. Models that explicitly preserve modality boundaries—MMGCN, MMGAT, and NSG-MoE—substantially outperform approaches such as UniGraph2 that concatenate multimodal features before propagation.
Early fusion tends to entangle heterogeneous feature distributions, making it difficult for GNNs to capture asymmetric multimodal semantics, whereas NSG’s structured separation avoids this collapse.
(3) Model performance is also influenced by the choice of pretrained modality encoders. Multimodal large language models, with stronger joint alignment and larger parameterization, generally produce higher-quality initial features, leading to superior downstream performance even for plain GNN baselines. While models using internally aligned MLLMs benefit from semantically consistent embeddings across modalities, our NSG-MoE remains competitive even with smaller encoders or external alignment (Table \ref{tab:magb}). This robustness stems from NSG-MoE’s augmented graph structure and its ability to route information through expert GNNs specialized for different structural patterns.
(4) Furthermore, both NSG-MoE and MMGAT consistently outperform UniGraph2, illustrating the effectiveness of GNN-based expert architectures. While conventional MoE modules excel at high-dimensional feature extraction, graph-structured MoE components capture relational patterns that standard MLP-based experts cannot. MMGAT benefits from its modality-aware attention, but its performance is constrained by the homogeneous expert architecture and its relatively coarse integration of modalities. NSG-MoE, in contrast, leverages multiple heterogeneous experts that specialize to different NSG configurations.

Results for link prediction, shown in Table \ref{tab:lp}, follow similar trends. Models generally perform better when using ImageBind features compared to CLIP, reflecting stronger alignment between visual and textual semantics. NSG-MoE again outperforms all baselines across both encoders. A noteworthy observation is that MMGCN occasionally underperforms the base GNN models, which can be attributed to its complete separation of modalities during message passing that limits cross-modal signal flow.

\begin{figure}[t]
    \centering
    \begin{subfigure}[b]{0.6\linewidth} 
        \centering
        \includegraphics[width=\textwidth]{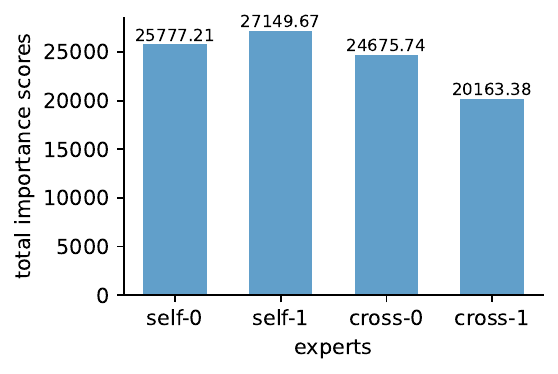} 
        \caption{Modality 0 (textual)}
    \end{subfigure}
    \hfill 
    \begin{subfigure}[b]{0.6\linewidth}
        \centering
        \includegraphics[width=\textwidth]{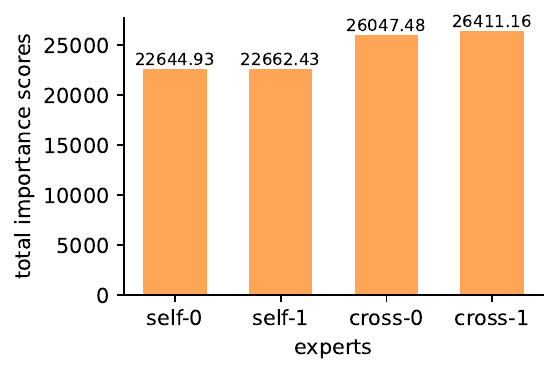} 
        \caption{Modality 1 (visual)}
    \end{subfigure}
    \caption{Analysis on the gating behavior. (node classification on Ele-fashion with CLIP; The same applies unless otherwise specified.)}
    \label{fig:gating}
\end{figure}
\textbf{Gating behavior.}
To better understand how NSG-MoE leverages its experts, we examine the gating patterns (Figure \ref{fig:gating}). The gating function exhibits clear specialization across experts. For text-dominant nodes, the model preferentially activates experts operating on self-type connections, indicating that textual semantics benefit from strong intra-modal consistency. Within the same expert type, importance varies markedly across experts, confirming that multiple experts per structural configuration successfully capture different input distributions.
For the visual modality, the relative importance across experts of the same type is more uniform. Nonetheless, image-based nodes show a consistent preference for experts processing cross-type edges, suggesting that visual information is enriched more effectively through textual descriptions of neighboring nodes than through purely visual neighbors. This aligns with our intuition that cross-type edges (effectively 2-hop neighbors over the self-graph) provide semantically complementary context.

\begin{figure}[t]
    \centering
    \includegraphics[width=\linewidth]{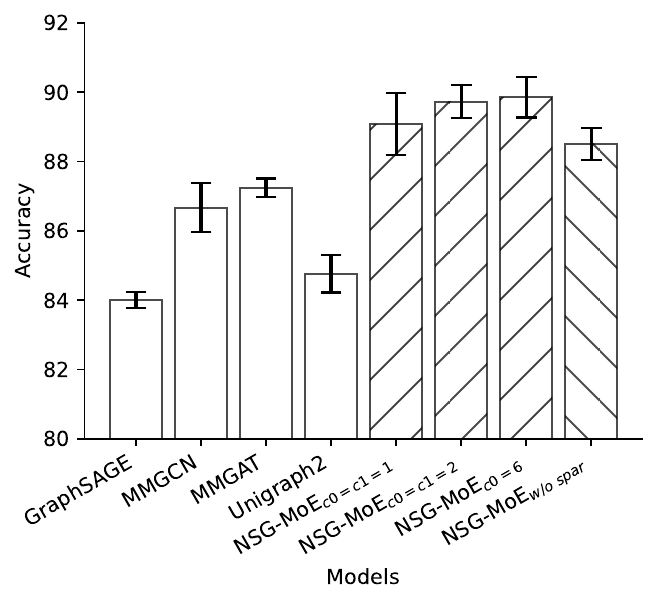}
    \caption{Results on ABIDE dataset. \(c_0=c_1=1\) represents that the intra-node connections are purely randomly sampled.}
    \label{fig:abide}
\end{figure}

\textbf{Effectiveness of sparsification for multi-modality graphs.}
We further evaluate NSG-MoE on datasets with more than two modalities, focusing on ABIDE to study the impact of sparsification. As shown in Figure \ref{fig:abide}, all NSG-MoE variants outperform existing baselines, with the configuration using global MST (\(c_0=6\)) achieving the best performance. The performance slightly declined when using approximate-MST as the generated tree may omit some strongly related connections.
Removing the MST-based sparsification module leads to a clear drop in performance, confirming that sparsification is crucial for preventing excessive cross-modal entanglement and for maintaining the structural interpretability of the augmented graph.

\begin{table}[t] \small
\centering
\caption{Ablation study. $*$ indicates discarding MoE (use only \(\Phi_{\text{intra}}\cup\Phi^\circ\)).}
\label{tab:abl}
\begin{tabular}{cccc}
\toprule
\multirow{2}{*}{\begin{tabular}[c]{@{}c@{}}Module\\ Implementation\end{tabular}} & nc & \multicolumn{2}{c}{lp} \\ \cmidrule(lr){2-2} \cmidrule(lr){3-4}
 & Ele-fashion & Sports & Cloth \\ \midrule
SAGE-h & 87.59 ±0.09 & 29.33 ±0.11 & 21.70 ±0.10 \\
HAN & 87.15 ±0.11 & 29.64 ±0.11 & 22.31 ±0.08 \\
HGT & 87.40 ±0.24 & 29.97 ±0.18 & 22.64 ±0.13 \\ \midrule
SAGE-h$^*$ & 87.04 ±0.23 & 28.09 ±0.21 & 21.19 ±0.08 \\
HAN$^*$ & 86.44 ±0.06 & 29.18 ±0.18 & 21.94 ±0.07 \\
HGT$^*$ & 86.83 ±0.07 & 29.42 ±0.04 & 22.05 ±0.01
\\ \bottomrule
\end{tabular}
\end{table}
\subsection{Ablation Study and Analysis}
Here we examine the contribution of the MoE module and the effect of different HGNN backbones. In addition to HGT, we evaluate SAGE-h and HAN \citep{wang2019heterogeneous} to understand how NSG-MoE behaves with different relational inductive biases. Results are summarized in Table \ref{tab:abl}.
The choice of base encoder significantly influences overall performance. For example, SAGE-h yields stronger results on Ele-Fashion but is noticeably weaker on the remaining datasets, reflecting the sensitivity of different HGNN designs to distinct graph structures and tasks.
Despite this variability, NSG-MoE consistently surpasses the basic models. Even if the backbone model is weak in a certain scenario, NSG-MoE still matches or exceeds prior state-of-the-art approaches based on GAT-style encoders—models generally regarded as strong default choices for graph data.
These observations highlight that the gains provided by NSG-MoE stem primarily from the proposed learning architecture, rather than from any unique advantage of a particular backbone.
(2) Removing the MoE component (the \emph{w/o MoE} variant) while retaining the NSG structure still leads to clear performance improvements over baseline models. This demonstrates that even without expert specialization, the combination of node splitting and heterogeneous message passing offers substantial benefits.


\begin{table}[t] 
\caption{Performance with different number of experts.}
\label{tab:expe}
\centering
\begin{tabular}{cccc}
\toprule
(\(n_1\),\(n_2\),\(k_e\)) & Ele-fashion & Sports & Cloth \\ \midrule
$(1,1,1)$ & 86.52 ±0.07 & 29.19 ±0.17 & 22.17 ±0.14 \\
$(2,2,1)$ & 87.13 ±0.04 & 29.83 ±0.09 & 22.23 ±0.12 \\
$(2,2,2)$ & 87.40 ±0.24 & 29.97 ±0.18 & 22.64 ±0.13 \\
$(2,2,4)$ & 86.89 ±0.11 & 29.65 ±0.12 & 22.26 ±0.07 \\ \bottomrule
\end{tabular}
\end{table}

\begin{figure}[t]
    \centering
    \includegraphics[width=\linewidth]{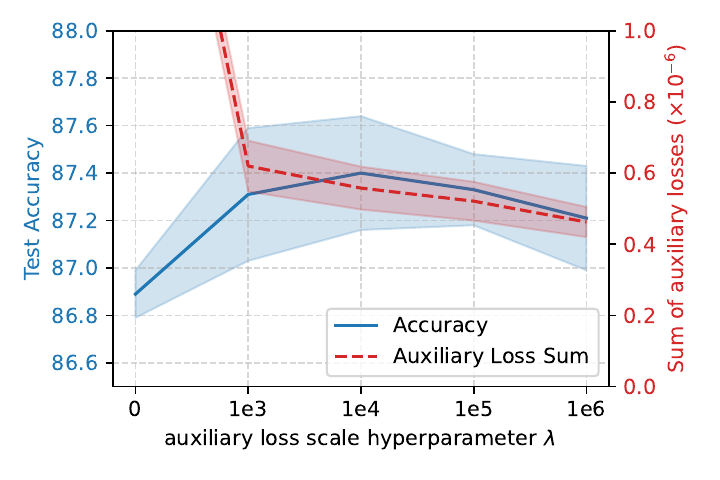}
    \caption{Effect of auxiliary loss scale hyperparameters}
    \label{fig:hyper}
\end{figure}
\subsection{Parameter Sensitivity Analysis}
We next analyze the influence of key hyperparameters in the MoE module. Table \ref{tab:expe} presents results under varying numbers of experts and different values of \(k_e\), the number of experts selected during routing.
When \(n_1=n_2=1\), the performance is close to that of a single expert, as it cannot fully leverage the advantages of the MoE architecture.
When \(n_1=n_2=2\), sparse expert selection (\(k_e<n_1+n_2\)) generally yields higher accuracy than dense routing, as dense routing tends to blur expert boundaries and introduces noisy or irrelevant signals for nodes, thus hindering the generalization.
However, when \(k_e\) becomes too small, e.g. \(k_e=1\), the model loses the ability to integrate complementary sources of information and the performance drops accordingly.
The optimal configuration across datasets is obtained with a larger number of experts and a moderate number of active experts, which provides enough diversity for specialization without overwhelming the routing mechanism.
Besides, performance decreases when removing either group of experts (either \(n_1\) or \(n_2\) is set to 0, as shown in Table \ref{tab:pre}). This result reinforces the importance of jointly modeling both self-type and cross-type structural patterns.

We further examine sensitivity with respect to $\lambda$, which controls the influence of the load-balancing and importance-regularization losses (Figure \ref{fig:hyper}). When $\lambda$ approaches zero, the model collapses towards using several specific experts, leading to severe imbalance and degraded performance. Increasing $\lambda$ initially helps distribute assignments more evenly, reducing auxiliary losses and improving accuracy. The best results occur at $\lambda = 10^4$; larger values over-regularize the routing network, causing the auxiliary losses to dominate the optimization and reduce accuracy. These findings emphasize the critical role of load-balancing in ensuring stable and meaningful expert specialization.


\begin{table}[]
\caption{Averaged training time per epoch (ms).}
\label{tab:eff}
\centering
\begin{tabular}{cccc}
\toprule
Model & Ele-fashion & Sports & Cloth \\ \midrule
SAGE & 13 & 15 & 22 \\
MMGCN & 27 & 42 & 69 \\
Unigraph2 & 224 & 317 & 503 \\
MMGAT & 108 & 170 & 281 \\
NSG & 28 & 46 & 74 \\
NSG-MoE & 106 & 177 & 275
\\ \bottomrule
\end{tabular}
\end{table}
\subsection{Efficiency Analysis}
We compare the training-time efficiency of NSG-MoE with several baselines, averaging results over ten runs to minimize stochastic variation. Table \ref{tab:eff} presents per-epoch training times.
The node splitting step is performed once during preprocessing and therefore does not influence per-epoch runtime. Compared with naïvely dividing modalities and propagating independently, our NSG construction introduces negligible overhead relative to standard message passing (as reflected by the similar running times of MMGCN).
Introducing a MoE module results in a total cost that is approximately equal to the original cost multiplied by the number of experts.
Further parallelizing the forward processes of each expert could potentially improve efficiency and enable the handling of larger-scale graphs in real-world applications.
UniGraph2 is noticeably slower due to its dependence on expensive pairwise shortest-path computations. MMGAT achieves similar efficiency to NSG-MoE, as both operate with identical numbers of total and active experts.
Overall, the results demonstrate that NSG-MoE balances expressive power and computational efficiency, making it well-suited for large-scale multimodal graph applications where both structural resolution and runtime constraints are critical.

\section{Conclusion}
In this work, we introduce NSG-MoE, a principled approach for multimodal graph learning that integrates a modality-aware node-splitting mechanism with a Mixture-of-Experts architecture.
By separating node representations into structured subcomponents and routing heterogeneous relations to specialized experts, the proposed framework effectively resolves the cross-modal entanglement issue and prevents noisy over-mixing of incompatible neighbors.
Our empirical study demonstrates the practical advantages of this design. Across diverse datasets, NSG-MoE consistently outperforms both classical GNNs and more sophisticated multimodal graph models. Moreover, the training-time comparison shows that NSG-MoE maintains competitive efficiency.
Beyond empirical performance, our theoretical analysis offers an understanding of the filtering behavior of NSG and why it generalizes well.
Future work may explore extending the expert routing to dynamic or task-adaptive settings, or incorporating temporal modalities.

\bibliographystyle{elsarticle-harv} 
\bibliography{mmgraph}

\end{document}